\documentclass[conference]{IEEEtran}
\IEEEoverridecommandlockouts 

\usepackage{amsmath,amssymb,amsthm}
\usepackage{mathtools}
\usepackage{graphicx}
\usepackage{booktabs}
\usepackage{array}
\usepackage{xcolor}
\usepackage{hyperref}
\usepackage{cite}
\usepackage{algorithm}
\usepackage[noend]{algpseudocode}

\hypersetup{colorlinks=false}

\newcommand{\figsep}{\vspace{-14pt}}

\newcommand{\bel}{b}
\newcommand{\spc}{\mathcal{S}}
\newcommand{\Ent}{\mathrm{H}}

\newcommand{\Istep}{I_{\mathrm{step}}}
\newcommand{\bprior}{b_{\mathrm{prior}}}
\newcommand{\bvisit}{b_{\mathrm{visit}}}
\newcommand{\bmerged}{b_{\mathrm{merged}}}
\newcommand{\Hrat}{H_{\mathrm{ratio}}}

\newtheorem{problem}{Problem}
\newtheorem{assumption}{Assumption}

\begin{document}\title{Entropy-Gated Belief Coordination for Decentralized Multi-Agent Search Under Intermittent Communication}

\author{%
  \IEEEauthorblockN{Mohamed Abdelnaby, Kevin Leahy}
  \IEEEauthorblockA{Department of Robotics Engineering,
    Worcester Polytechnic Institute\\
    \texttt{\{mabdelnaby,\,kleahy\}@wpi.edu}
    \thanks{This work was supported by ONR under grant N00014-25-1-2225.}}
}

\maketitle

\begin{abstract}
We study decentralized multi-agent target search where homogeneous agents communicate intermittently at Poisson-distributed times. Standard unconditional belief fusion wastes communication opportunities by synchronizing agents during high-entropy exploration, when diverse independent beliefs provide better coverage than a premature consensus.
We introduce \emph{entropy-gated belief coordination}, in which agents skip fusion while their collective entropy ratio exceeds a threshold~\(\theta\) and merge only during exploitation, consistent with the bifurcation structure of nonlinear opinion dynamics and the submodular structure of the per-step information gain objective. 
We further derive \(\Istep(\alpha,\beta)\), the expected mutual information per observation step between two agents' binary sensors, as an interpretable, communication-free measure of sensor informativeness that motivates the gating design and guides system-level analysis. 
Experiments across 103{,}680 trials (nine grid sizes up to \(100{\times}100\), Poisson communication timing, four target movement patterns) show that the Entropy-Gated Trust-Decay Planner (\textsc{EG-TDP}), which adds a detection-probability planner switch in exploitation mode, achieves mean belief quality \(\bar{Q}=0.300\) (mean belief mass at the true target cell, averaged across all trials and steps), a \(58.9\%\) gain over arithmetic mean and a \(37.4\%\) gain over visit-weighted fusion. On representative configurations, EG-TDP also outperforms a joint-Bayesian reference that uses all agents'~observations at every step, despite operating under random intermittent contact only.
\end{abstract}

\section{Introduction}\label{sec:intro}

Coordinated multi-agent search arises in search-and-rescue, environmental monitoring, and inspection, where a team must locate a target of interest with noisy sensors and intermittent communication. Each agent maintains a probabilistic belief map updated via Bayesian recursion from local observations. At communication events, agents fuse beliefs to pool independently gathered evidence.

The Kullback-Leibler (KL) divergence provides a principled framework to decide how to combine their beliefs into a single consensus map when agents meet. The minimizer of a weighted forward KL objective is the arithmetic mean and the minimizer of the reverse KL is the geometric mean~\cite{genest1986,stone1961}. Our prior work~\cite{cdc2025} applied these classical results to multi-agent tracking and introduced visit-weighted fusion to suppress ghost targets, with an implementation linear in the number of agents and the size of the environment. The present paper asks instead \emph{when} agents should fuse, and how the planner should respond to the team's information state.

Two problems motivate this work. First, fusing beliefs during high-entropy exploration synchronizes agents who should be maintaining diverse, complementary views of the search space. Premature consensus forces all agents to plan from the same belief, causing them to select correlated positions and undermining their search efforts. Second, a natural response to stale beliefs from long communication blackouts is \emph{temporal trust decay}, in which the merged belief is weighted toward a prior or uniform distribution in proportion to elapsed blackout time on the grounds that older evidence is less reliable. However, the decay factor cannot distinguish between a correct peak and a ghost peak, because both produce the same entropy. Therefore, this method cannot be applied directly to the belief map.

In this work, we
\begin{enumerate}
  \item derive \(\Istep(\alpha,\beta)\), a communication-free constant
    computable from the sensor false-positive rate \(\alpha\) and false-negative rate \(\beta\) alone, that quantifies how quickly two homogenous agents' beliefs diverge during a communication blackout and informs the design of the coordination policy;
  \item introduce entropy-gated merging (\textbf{EG-KL}), in which agents fuse beliefs only when team entropy falls below threshold~\(\theta\), preserving exploration diversity and decoupling performance from communication frequency;
  \item propose \textbf{EG-TDP}, which adds a planner switch from
    entropy-reduction to detection-probability scoring upon entering
    exploitation mode, enabling convergence at large scales where
    entropy-reduction Model Predictive Control (MPC) alone cannot commit
    to pursuit within the step budget, while reducing inter-agent belief transmission by over 97\% at $100{\times}100$ scale; and
  \item propose \textbf{EG-Istep}, a variant of EG-TDP that replaces
    the binary planner switch with a continuous blend weighted by
    \(\gamma\in(0,1]\), a trust-decay factor derived from \(\Istep\) that
    decays with blackout length to discount detection-probability scoring
    as beliefs grow stale, providing a sensor-dependent criterion for
    choosing between continuous and hard-threshold planner switching.
\end{enumerate}

\section{Related Work}
 
Our work sits at the intersection of distributed belief fusion, multi-robot search under communication constraints, and coordination under intermittent connectivity. We describe each area and the specific gap that motivates our approach.
 
A substantial body of work addresses how agents should combine probability distributions when they meet. Classical opinion pooling theory establishes that logarithmic pooling (geometric mean) is externally Bayesian when agents share a likelihood model, while linear pooling (arithmetic mean) is preferred under potential bias~\cite{genest1986,stone1961,clemen1999}. Covariance intersection~\cite{julier1997} provides conservative Gaussian fusion guarantees under unknown correlations, and event-triggered schemes~\cite{trimpe2014,dimarogonas2012} reduce communication overhead while maintaining stability. None of these works poses the conditional question of whether agents should adopt a shared belief at all on a given event, or how that belief should alter path planning. Our prior work~\cite{cdc2025} derived KL-optimal pooling rules and visit-weighted variants for binary-sensor tracking but applied fusion unconditionally, leaving open the question of when fusion is beneficial and how the planner should respond.
 
Several approaches address multi-robot active information gathering under limited communication. Lauri et al.~\cite{lauri2020} formulate the problem as a Dec-POMDP but require complete knowledge of target dynamics and treat communication as absent rather than Poisson-distributed. Deep reinforcement learning approaches such as~\cite{zhou2023} learn cooperative strategies end-to-end but are sensitive to the training environment and do not address when to share beliefs. Krause et al.~\cite{krause2008} and Corah and Michael~\cite{corah2019} establish near-optimal greedy strategies under submodular information objectives. Earlier treatments of decentralized probabilistic search established the problem structure underlying this literature~\cite{bourgault2003,chungburdick2012,chung2011}. Across this literature, agents share information freely and the shared representation is immediately used for planning.
 
The work most directly related to ours is Ristic and Skvortsov~\cite{ristic2020}, who study decentralized area-restricted search with intermittent information and consensus coordination. When agents communicate they always fuse. Hollinger et al.~\cite{hollinger2015} address \emph{how} to fuse during multirobot search by weighting observations by communication channel quality, but the fusion decision is treated as fixed whenever agents meet. Roth et al.~\cite{roth2006} adopt a multi-agent POMDP framework with stronger optimality guarantees, but joint policy computation becomes intractable at the grid scales we consider. The question of whether to fuse is never raised, and the planner objective does not change with collective confidence. Event-triggered control~\cite{trimpe2014,dimarogonas2012} determines when to communicate but treats the fusion rule and its planning consequence as fixed. Our work separates these decisions. Given that a Poisson event has fired, should the team merge beliefs, and how should that change the way agents plan?
 
Deciding when to merge beliefs and how to plan with the result is a form of the exploration-exploitation tradeoff. Bizyaeva et al.~\cite{bizyaeva2023} formalize this as a bifurcation between a disagreement regime, where agents maintain diverse independent views, and an agreement regime, where they commit to a shared estimate, controlled by a sensitivity parameter. This motivates using the team's collective entropy as a coordination signal, formalised in Section~\ref{sec:gate}. The opinion dynamics model operates on continuous scalar opinions with no planning component. Translating the bifurcation insight into an operational policy for discrete belief maps, coupled with a planning objective that adapts when the regime switches, is the contribution of this paper.

\section{Problem Formulation}

\subsection{Environment and Sensors}

We consider a team of $N$ agents moving in a discrete environment
and searching for a single target. The environment is modeled as
a grid $\spc$ of $|\spc|=R{\times}C$ cells.
Agent $i\in\{1,\ldots,N\}$ occupies cell $x_{i,t}\in\spc$ at time $t$,
the target occupies $x^*_t\in\spc$.
Each agent carries a binary sensor with false positive rate \(\alpha\)
and false negative rate \(\beta\). At each time step, agent \(i\)
produces a binary observation \(z_{i,t}\in\{0,1\}\) of its current
cell, where \(z_{i,t}=1\) indicates a putative detection. The
observation probability is:
\begin{equation}
  P(z_{i,t}\!=\!1\mid x^*_t,x_{i,t}) =
  \begin{cases} 1{-}\beta & x^*_t = x_{i,t} \\ \alpha & \text{otherwise.}
  \end{cases}
  \label{eq:sensor}
\end{equation}
Agent \(i\)'s belief \(\bel_{i,t}:\spc\to[0,1]\) evolves by Bayesian
recursion:
\begin{equation}
  \bel_{i,t+1}(s) = \eta_i\cdot P(z_{i,t}\mid s,x_{i,t})\cdot\bel_{i,t}(s),
  \label{eq:bayes}
\end{equation}
where \(\eta_i\) is a normalization constant ensuring \(\sum_s \bel_{i,t+1}(s)=1\). Equation~\eqref{eq:bayes} contains no prediction step and therefore treats the target as stationary. Any diffusion kernel would be an additional parameter, and we choose not to assume any knowledge about the target's model. The effect of this assumption on faster-moving targets is examined in Section~\ref{sec:results_pattern}.
All agents start with a Gaussian prior centered on the last-known
target location \((r_0,c_0)\):
\begin{equation}
  \bel_{i,0}(s)\propto
  \exp\!\Bigl(-\tfrac{\|(r_s,c_s)-(r_0,c_0)\|^2}{2\sigma_0^2}\Bigr),
  \label{eq:prior}
\end{equation}
where $\sigma_0$ controls how diffuse the initial belief is.

\subsection{Communication Model}

Communication events arrive as a Poisson process with mean inter-event duration \(\bar{k}\). This models scenarios in which communication windows open unpredictably, for instance when agents pass within radio range of each other or atmospheric conditions permit a brief link. At each event all \(N\) agents communicate simultaneously. Since the events are Poisson, the blackout duration between consecutive events follows an exponential distribution with mean \(\bar{k}\), its realization is unknown to agents in advance.

\subsection{Objective}

We measure performance by how well the collective belief
localizes the target throughout the search. The instantaneous belief
quality at step $t$ is the mean probability mass assigned to the true
target cell across all agents:
\begin{equation}
  Q_t = \frac{1}{N}\sum_{i=1}^N \bel_{i,t}(x^*_t).
  \label{eq:quality}
\end{equation}
High $Q_t$ indicates that the team's evidence points toward the correct
location, and low $Q_t$ reflects either diffuse beliefs (little information
gathered) or misdirected beliefs (ghost peaks). A \emph{trial} is a
single simulation run of at most $T_{\max}$ steps from a fixed initial configuration. For a single trial $\tau$, the per-trial quality is
$\widehat{Q}^\tau=\frac{1}{T_{\max}}\sum_{t=1}^{T_{\max}} Q_t$, where the superscript $\tau$ identifies the trial instance. The reported metric is the mean over all trials $\mathcal{T}$, including those in which the target is never found:
\begin{equation}
  \bar{Q} = \frac{1}{|\mathcal{T}|}\sum_{\tau\in\mathcal{T}}\widehat{Q}^\tau.
  \label{eq:qbar}
\end{equation}
Because $Q_t$ measures belief mass at the true target cell, it is naturally low throughout any trial in which the team fails to localise the target. Importantly, $\bar{Q}$ also distinguishes between methods that succeed by directed pursuit versus coverage luck. A method may stumble upon the target with a diffuse belief, registering a success but keeping $Q_t$ near zero throughout, while a method that concentrates beliefs efficiently scores high $Q_t$ during directed pursuit. $\bar{Q}$ is therefore more sensitive than binary success rate to whether the coordination policy is working as intended.

A coordination policy $\pi = (\pi_\mathrm{fuse}, \pi_\mathrm{plan})$ operates on two timescales. $\pi_\mathrm{fuse}: \Hrat \to \{0,1\}$ maps the team entropy ratio at each communication event to a binary merge decision. $\pi_\mathrm{plan}: \{\bel_{i,t}\}_{i=1}^N \to \spc^N$ maps current local beliefs to a joint action at each time step.

 \begin{assumption}\label{ass:info}
Each agent observes its own local belief $\bel_{i,t}$, sensor parameters $(\alpha,\beta)$, and the number of steps $k$ elapsed since the last communication event. At communication events, agents exchange scalar entropy values $\mathrm{H}(\bel_i)$ to compute $\Hrat$ jointly. Agents do not have access to true target location $x^*_t$, belief quality $Q_t$, or the timing of future events.
\end{assumption}

\begin{problem}
Given $N$ agents with sensor parameters $(\alpha,\beta)$, a Poisson communication process with mean inter-event duration $\bar{k}$, and a Gaussian prior over the target's initial location, find a policy $\pi$ subject to Assumption~\ref{ass:info} that maximises $\bar{Q}$.
\end{problem}

\section{Technical Approach}

\begin{figure}[t]
  \centering
  \includegraphics[width=\columnwidth]{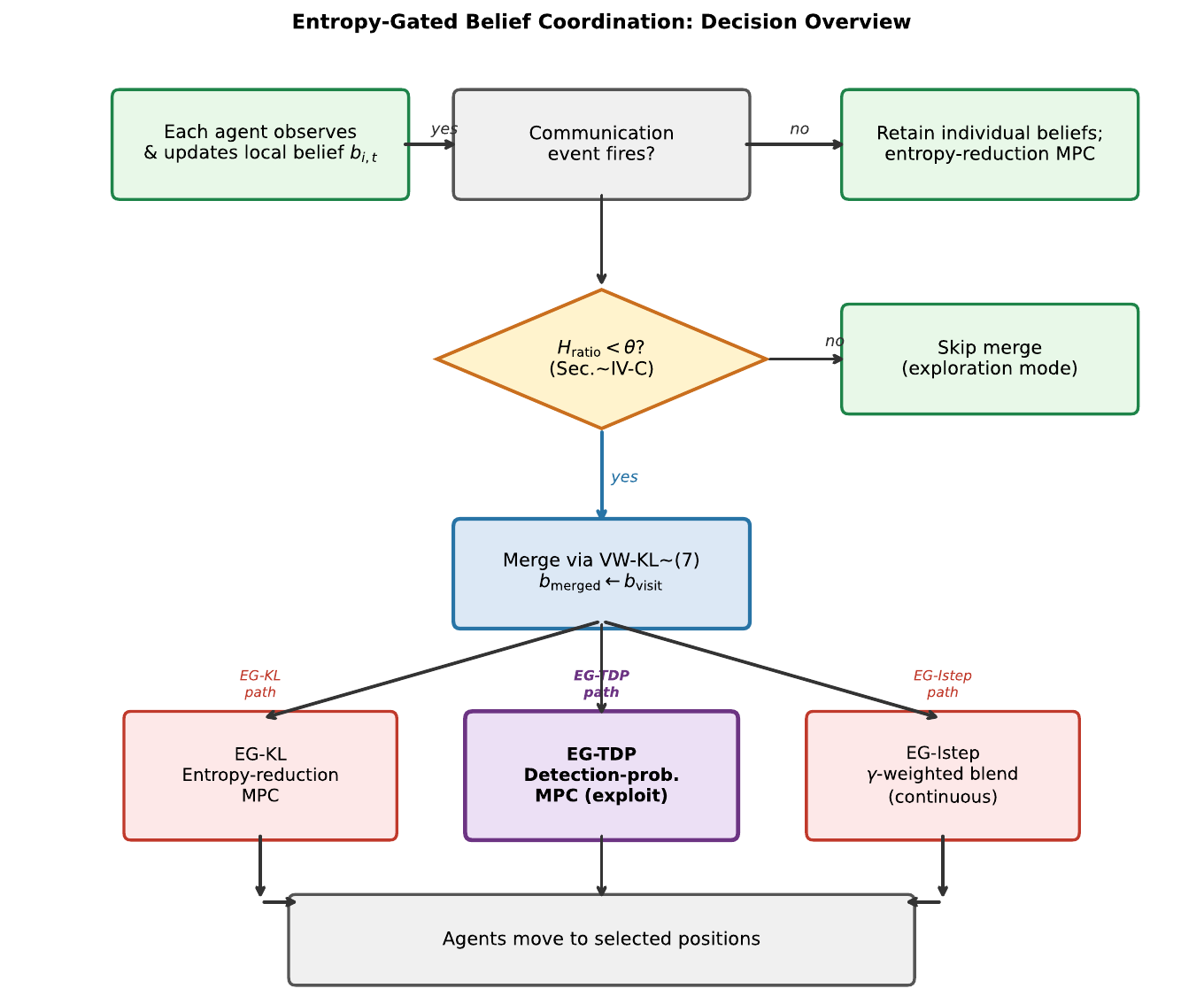}
  \figsep
  \caption{Decision overview for the three proposed coordination policies. At each communication event the team computes $\Hrat$, the entropy gate determines whether to merge and, for EG-TDP and EG-Istep, which planner objective to use.}
  \label{fig:overview}
\end{figure}

Effective multi-agent search requires two distinct coordination modes. During \emph{exploration}, agents' beliefs are still diffuse and each agent should independently gather information from different regions of the grid. Forcing a shared belief at this stage causes agents to plan correlated paths and waste coverage. During \emph{exploitation}, beliefs have concentrated sufficiently that coordinating toward the most probable target location is more valuable than continued independent search. Our method detects this transition from a scalar summary of the team's collective uncertainty, the entropy ratio $\Hrat$, defined in Section~\ref{sec:gate}, and uses a threshold $\theta$ to decide when to switch modes. Early in a search the team's beliefs are diffuse and $\Hrat$ is near its maximum, as agents gather evidence and beliefs concentrate, $\Hrat$ falls. Once it crosses below $\theta$ a communication event triggers a belief merge (Section~\ref{sec:vwkl}) and, for \textsc{EG-TDP}, a switch to the detection-probability planner (Section~\ref{sec:egtdp}). Figure~\ref{fig:overview} summarises the full decision structure across all proposed methods.

The following subsections present the base merge formula, the per-step information-loss characterization, and the three proposed coordination policies (EG-KL, EG-TDP, and EG-Istep).

\subsection{Visit-Weighted Belief Merging}\label{sec:vwkl}

The merging formula from~\cite{cdc2025} serves as the base rule applied
during exploitation.
Each cell \(s\) receives a weight proportional to the number of times agent \(i\) has visited it since the last merge, denoted \(v_i(s)\):
\begin{equation}
  w_i(s)=\frac{v_i(s)+1}{\sum_j(v_j(s)+1)},
  \label{eq:weights}
\end{equation}
where the \(+1\) prevents division by zero for cells that
agent $i$ has not visited since the last merge ($v_i(s)=0$).
The fused belief is:
\begin{equation}
  \bvisit(s)=\sum_{i=1}^N w_i(s)\,\bel_i(s).
  \label{eq:visitmerge}
\end{equation}
Visit counts reset after each merge to prevent double-counting.

\subsection{Belief Divergence During Blackouts}

During a communication blackout each agent independently updates its local belief from its own observations. Without sharing these updates, agents searching different parts of the grid can develop very different views of where the target is, even though both are reasoning correctly from their respective evidence. The longer the blackout, the more these individual beliefs can diverge. This section derives a single sensor-level constant that quantifies how fast that divergence happens, computable from the sensor parameters alone without any inter-agent communication.

Consider two agents each observing the same cell independently through sensor~\eqref{eq:sensor}. Let $p\in[0,1]$ be the prior probability that the target occupies that cell. Each agent produces a binary observation, and both have the same marginal detection probability $q(p)=\alpha+p(1{-}\alpha{-}\beta)$ obtained by marginalizing over whether the target is present. The agent detects with probability $p(1{-}\beta)$ (true positive) or $(1{-}p)\alpha$ (false positive).

The shared information between the two agents' observations, that is,
how much knowing one agent's result tells you about the other's, is
the mutual information $\mathrm{I}(z_A;z_B\mid p)$. Because $z_A$
and $z_B$ each have binary entropy $\mathrm{H}_2(q(p))$, by the
definition of mutual information:
\begin{equation}
  \mathrm{I}(z_A;z_B\mid p)=2\,\mathrm{H}_2(q(p))-\mathrm{H}(z_A,z_B\mid p),
  \label{eq:MIp}
\end{equation}
where $\mathrm{H}(z_A,z_B\mid p)$ is the joint entropy of both
observations. Because $z_A$ and $z_B$ are conditionally independent given the target state, the joint distribution is the following $2{\times}2$ table:
\begin{center}\small
\begin{tabular}{@{}c|cc@{}}
 & $z_B{=}1$ & $z_B{=}0$ \\ \hline
$z_A{=}1$ & $p(1{-}\beta)^2{+}(1{-}p)\alpha^2$ & $p(1{-}\beta)\beta{+}(1{-}p)\alpha(1{-}\alpha)$ \\
$z_A{=}0$ & $p(1{-}\beta)\beta{+}(1{-}p)\alpha(1{-}\alpha)$ & $p\beta^2{+}(1{-}p)(1{-}\alpha)^2$ \\
\end{tabular}
\end{center}

Since we do not know $p$ in advance, we average over the uniform
prior on $[0,1]$, giving a sensor-level constant:
\begin{equation}
  \boxed{\Istep(\alpha,\beta)=\int_0^1\mathrm{I}(z_A;z_B\mid p)\,\mathrm{d}p.}
  \label{eq:Istep}
\end{equation}
The joint table above makes $\mathrm{H}(z_A,z_B\mid p)$ closed-form for any $(\alpha,\beta,p)$. In practice we approximate the integral by a Riemann sum over a uniform discretization of $p\in[0,1]$, which converges rapidly since the integrand is smooth. $\Istep$ depends only on $\alpha$ and $\beta$ and requires no
communication between agents to compute. Sensors with higher $\Istep$
share more information per observation step, equivalently, their
independent beliefs diverge faster during a blackout, making coordination
more valuable.

\begin{figure}[t]
  \centering
  \includegraphics[width=\columnwidth]{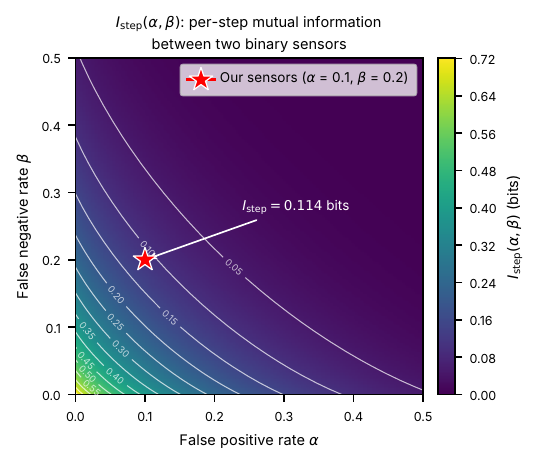}
  \figsep
  \caption{$\Istep(\alpha,\beta)$ over the binary sensor parameter space. Higher values indicate faster belief divergence during blackouts, making coordination more valuable. The star marks our sensor profile ($\alpha{=}0.1,\beta{=}0.2$, $\Istep{=}0.114$~bits).}
  \label{fig:istep}
\end{figure}

Figure~\ref{fig:istep} shows $\Istep(\alpha,\beta)$ over the full sensor parameter space. $\Istep$ is highest 0.7213 for informative sensors (low $\alpha,\beta$) and zero for random sensors ($\alpha{=}\beta{=}0.5$), where detections carry no information about target presence. Our sensors ($\alpha{=}0.1,\beta{=}0.2$) give $\Istep{=}0.114$~bits, marked on the figure.

\textbf{Trust-decay factor.}
After $k$ steps of independent observation, with each step eroding
$\Istep$ bits of shared information, the accumulated divergence
suggests discounting the reliability of any merged belief from
that blackout. We capture this with:
\begin{equation}
  \gamma=\exp\!\bigl(-k\log_2 N\cdot\Istep(\alpha,\beta)\bigr),
  \label{eq:gamma}
\end{equation}
where $k$ is the number of steps since the last communication event,
$N$ is the team size, and $\gamma\in(0,1]$ is the resulting trust
weight. At $k=0$ (just communicated), $\gamma=1$ and the merged
belief is fully trusted. As $k$ grows, $\gamma$ decays toward zero,
reflecting that a belief formed from increasingly stale pairwise
evidence should carry less weight in the planner's decisions.

The $\log_2 N$ factor accounts for team size. Naively, one might scale by $(N{-}1)$, the number of agent pairs, but this overcounts. All $N$ agents are observing the same target, so their observations are correlated through that shared latent variable. Each successive agent contributes diminishing marginal information, $\log_2 N$ approximates this sublinear growth, where each doubling of team size adds one additional unit of coordination benefit rather than a full linear increment.

\subsection{Entropy-Gated Merging (EG-KL)}\label{sec:gate}

The central question is when agents should pool their beliefs. Fusing during exploration synchronizes agents who should cover complementary regions and reduces collective coverage, fusing during exploitation amplifies a shared signal and coordinates pursuit. This matches the disagreement--agreement bifurcation of nonlinear opinion dynamics~\cite{bizyaeva2023}.

The per-step information gain is monotone submodular in agent observation positions~\cite{krause2008,nemhauser1978}, so independent greedy planning is approximately optimal for single-step coverage. Premature consensus forces correlated position choices, reducing coverage toward what a single agent achieves. The entropy gate preserves the diversity that makes independent planning effective.

To detect which regime the team is currently in, we track the mean
per-agent entropy normalized by its maximum possible value:
\begin{equation}
  \Hrat=\frac{1}{N}\sum_{i=1}^N\Ent(\bel_i)\Big/\log_2|\spc|,
  \label{eq:hratio}
\end{equation}
computed from individual pre-merge beliefs.
$\Hrat\approx1$ indicates beliefs are nearly uniform (exploration).
$\Hrat\approx0$ indicates beliefs are sharply peaked (exploitation).

At each communication event, the team applies the following rule:
\begin{equation}
  \text{merge if and only if}\quad\Hrat<\theta.
  \label{eq:gate}
\end{equation}
When $\Hrat<\theta$ (exploitation), agents pool their beliefs via~\eqref{eq:visitmerge}, broadcast $\bmerged$, and reset visit counts. When $\Hrat\geq\theta$ (exploration), the merge is skipped and each agent continues from its own belief. The gate never blocks a communication event, it only determines whether beliefs are fused once agents are in contact.

\subsection{Entropy-Gated Trust-Decay Planner (EG-TDP)}\label{sec:egtdp}

\textsc{EG-TDP} augments the gated merge with a planner objective switch at the exploitation threshold. The rationale is as follows. During exploration ($\Hrat\geq\theta$), the team has not yet localized the target, the appropriate goal is to maximize information gain across the grid, and agents use standard entropy-reduction MPC independently. During exploitation ($\Hrat<\theta$), beliefs have concentrated and the priority shifts to directing agents toward the most probable target location.

Let $\mathbf{x}=(x_1,\ldots,x_N)$ denote a proposed joint action, where $x_i\in\spc$ is the cell agent $i$ would occupy at the next step.
From sensor model~\eqref{eq:sensor}, the probability that agent $i$ obtains a positive detection at $x_i$ is $\alpha+\bmerged(x_i)(1{-}\alpha{-}\beta)$, which follows from marginalizing over target presence, the probability is $(1{-}\beta)$ if the target occupies $x_i$ (weighted by $\bmerged(x_i)$) plus $\alpha$ otherwise.
Summing over all agents gives the expected total number of detections from joint action $\mathbf{x}$.
We subtract an overlap penalty to discourage all agents from converging on a single cell:
\begin{equation}
  \mathrm{score}(\mathbf{x})=
    \sum_{i=1}^N\bigl[\alpha+\bmerged(x_i)(1{-}\alpha{-}\beta)\bigr]
    -\lambda_{\mathrm{ov}}\,\phi(\mathbf{x}),
  \label{eq:exploit}
\end{equation}
where $\phi(\mathbf{x})$ counts agents assigned to the same proposed cell and $\lambda_{\mathrm{ov}}$ weights the penalty.
Without $\phi$, the optimal action would send every agent to the single highest-belief cell. The penalty spreads agents across the top candidates, increasing joint coverage of the high-probability region.
After a merge event in exploitation mode, all agents adopt $\bmerged$, so Eq.~\eqref{eq:exploit} uses a single shared belief for all $i$.

\subsection{EG-Istep: \texorpdfstring{\(\Istep\)}{I-step}-Modulated Planner}\label{sec:egistep}

\textsc{EG-Istep} shares the same entropy gate as \textsc{EG-KL} and \textsc{EG-TDP} but replaces the binary planner switch with a continuous blend modulated by the trust-decay factor \(\gamma\)~\eqref{eq:gamma}. After \(k\) steps without communication, actions are scored by:
\begin{equation}
  \mathrm{score}(\mathbf{x}) =
    \gamma\,\widehat{d}(\mathbf{x}) + (1{-}\gamma)\,\widehat{\Delta H}(\mathbf{x}),
  \label{eq:istep-blend}
\end{equation}
where \(\widehat{d}\) is the normalised detection-probability score from~\eqref{eq:exploit} and \(\widehat{\Delta H}\) is the expected entropy reduction normalised by \(\log_2|\spc|\). The detection score \(\widehat{d}\) depends on the pre-blackout merged belief \(\bmerged\), while \(\widehat{\Delta H}\) uses each agent's current local belief only. Thus \(\gamma\) controls how much the planner relies on \(\bmerged\). At \(k{=}0\), \(\gamma{=}1\) and the planner pursues the merged estimate, as \(k\) grows and \(\bmerged\) grows stale, \(\gamma\) decays and the planner shifts toward exploration to gather fresh evidence.
 
\begin{figure}[t]
  \centering
  \includegraphics[width=\columnwidth]{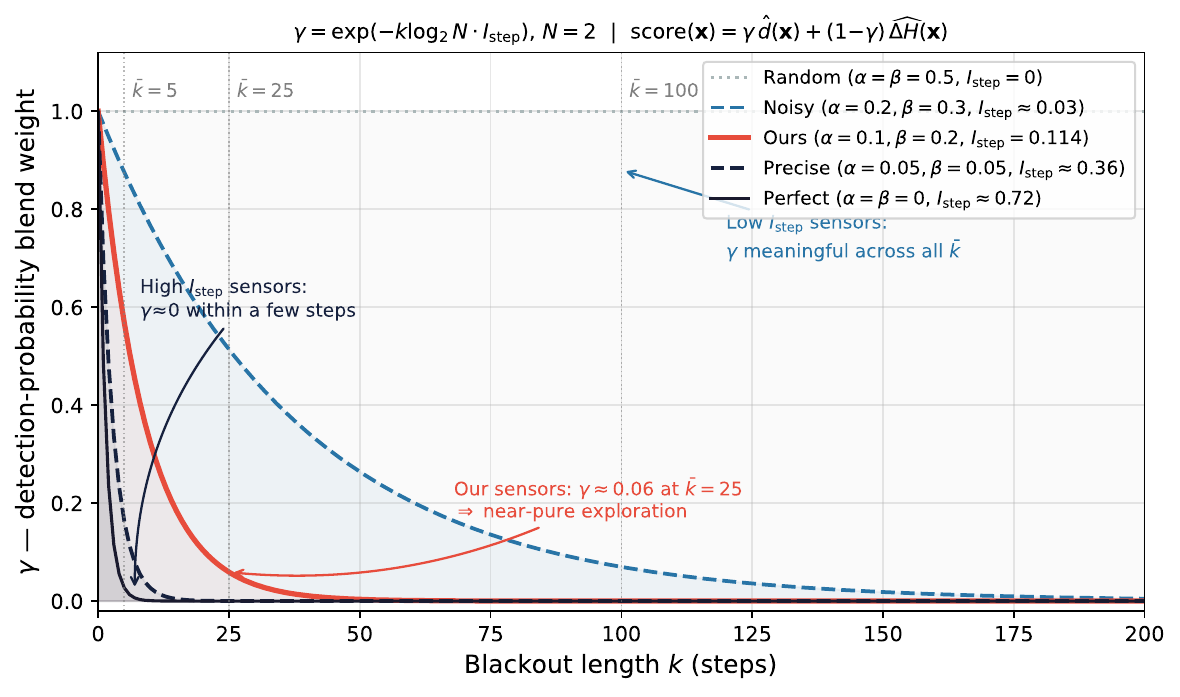}
  \figsep
  \caption{Trust-decay factor $\gamma$ as a function of blackout length $k$ for ($N{=}2$) agents. The complement $1{-}\gamma$ goes to entropy-reduction.}
  \label{fig:gamma_decay}
\end{figure}

Whether this continuous transition outperforms the binary switch of \textsc{EG-TDP} therefore depends on how quickly \(\bmerged\) becomes an unreliable guide for the planner. Figure~\ref{fig:gamma_decay} shows \(\gamma\) as a function of blackout length for representative sensor profiles. For low-\(\Istep\) sensors (noisy or near-random), each observation shifts the belief only slightly, so individual beliefs remain close to \(\bmerged\) throughout a typical blackout, \(\gamma\) stays above zero, the planner continues to weight detection scoring on \(\bmerged\), and the pre-blackout collective estimate remains a useful guide to target location. For high-\(\Istep\) sensors (informative, precise), a single observation can substantially reshape the belief map. After only a few steps the individual beliefs have moved away from \(\bmerged\), the pre-blackout estimate is stale, and $\gamma$ falls toward zero, shifting the planner entirely onto post-blackout entropy-reduction so agents gather fresh evidence before any collective estimate is reliable.
 
\begin{algorithm}[t]
\caption{Entropy-Gated Belief Coordination}\label{alg:framework}
\begin{algorithmic}[1]
\Require threshold \(\theta\), sensor \((\alpha,\beta)\), \(N\) agents
\State Initialise \(\bel_{i,0}\) from Gaussian prior~\eqref{eq:prior}
\For{each step \(t = 1, 2, \ldots\)}
  \State Each agent observes \(z_{i,t}\), updates \(\bel_{i,t}\)
    via~\eqref{eq:bayes}
  \If{communication event fires}
    \State Compute \(\Hrat\) via~\eqref{eq:hratio}; \(k\leftarrow\) steps since last event
    \If{\(\Hrat < \theta\)} \Comment{Exploitation: merge}
        \State \(\bmerged\leftarrow\bvisit\) via~\eqref{eq:visitmerge};
        broadcast to agents 
        \State reset \(v_i\)
    \Else \Comment{Exploration: skip merge}
      \State Agents retain individual beliefs
    \EndIf
  \EndIf
  \If{\textsc{EG-TDP} \textbf{and} \(\Hrat<\theta\)}
    \State Plan with detection-probability score~\eqref{eq:exploit}
  \Else
    \State Plan with entropy-reduction score
  \EndIf
  \State Agents move to selected positions
\EndFor
\end{algorithmic}
\end{algorithm}

Algorithm~\ref{alg:framework} presents the full coordination loop. Each iteration begins with agents observing and updating their beliefs (line~3). When a communication event fires (line~4), the team computes $\Hrat$ and either merges or defers (lines~5--9). When $\Hrat<\theta$, beliefs merge via~\eqref{eq:visitmerge} (line~7), otherwise each agent retains its own belief (line~9). EG-TDP additionally switches to the detection-probability planner~\eqref{eq:exploit} (line~13) when in exploitation, EG-KL always uses entropy-reduction planning (line~15).

\section{Experiments}

\subsection{Setup}

We evaluate on discrete grids ranging from \(10{\times}10\) to \(100{\times}100\) (nine sizes), with \(N\in\{2,3\}\) agents per trial. All agents use binary sensors with false positive rate \(\alpha{=}0.1\) and false negative rate \(\beta{=}0.2\), giving \(\Istep{=}0.114\)~bits. Communication follows a Poisson process with mean inter-event duration \(\bar{k}\) ranging over \(\{5,10,25,50,100,200\}\)~steps. We test four target movement patterns: stationary, random walk, evasive (moving away from the map centre), and patrol (a fixed corner-to-corner circuit). The Gaussian prior spread is \(\sigma_0{=}0.15\min(R,C)\), scaling with grid size so the initial belief covers a similar fraction of the search area regardless of environment scale. At each step the planner evaluates 50 randomly sampled joint actions with horizon \(H{=}3\) and selects the highest-scoring candidate. Each configuration is run for 30 trials with a step budget of 2500, yielding \(12{,}960\) trials per method and \(103{,}680\) total. The entropy threshold is \(\theta{=}0.6\) and the overlap penalty weight is \(\lambda_{\mathrm{ov}}{=}0.5\).

\subsection{Baselines and Comparisons}

The unconditional baselines merge at every communication event without any entropy check. \textsc{Arith} uses the arithmetic mean. \textsc{Geom} uses the geometric mean, which aggressively concentrates mass but can suppress recovery from incorrect peaks. \textsc{VW-KL} applies the visit-weighted formula~\eqref{eq:visitmerge} from~\cite{cdc2025} to reduce ghost targets. All three apply the same merge rule regardless of the team's current uncertainty.

\textbf{EG-KL} (ours) adds the entropy gate, applying \eqref{eq:visitmerge} only when \(\Hrat<\theta\), and agents plan independently with entropy-reduction MPC otherwise. \textbf{EG-TDP} (ours) additionally switches the planner to the detection-probability objective~\eqref{eq:exploit} in exploitation, directing agents toward the high-probability region rather than continuing broad coverage.

\textbf{EG-Istep} (ours) uses the same entropy gate as \textsc{EG-KL} and \textsc{EG-TDP} but replaces the binary planner switch with the continuous \(\gamma\)-weighted blend~\eqref{eq:istep-blend} described in Section~\ref{sec:egistep}.

The \textsc{Full-Comm} reference applies a joint Bayesian update at every step, it is included to provide a reference for the ceiling performance achievable with unconstrained communication.

\section{Results}

\subsection{Overall Belief Map Quality}

\begin{figure}[t]
  \centering
  \includegraphics[width=\columnwidth]{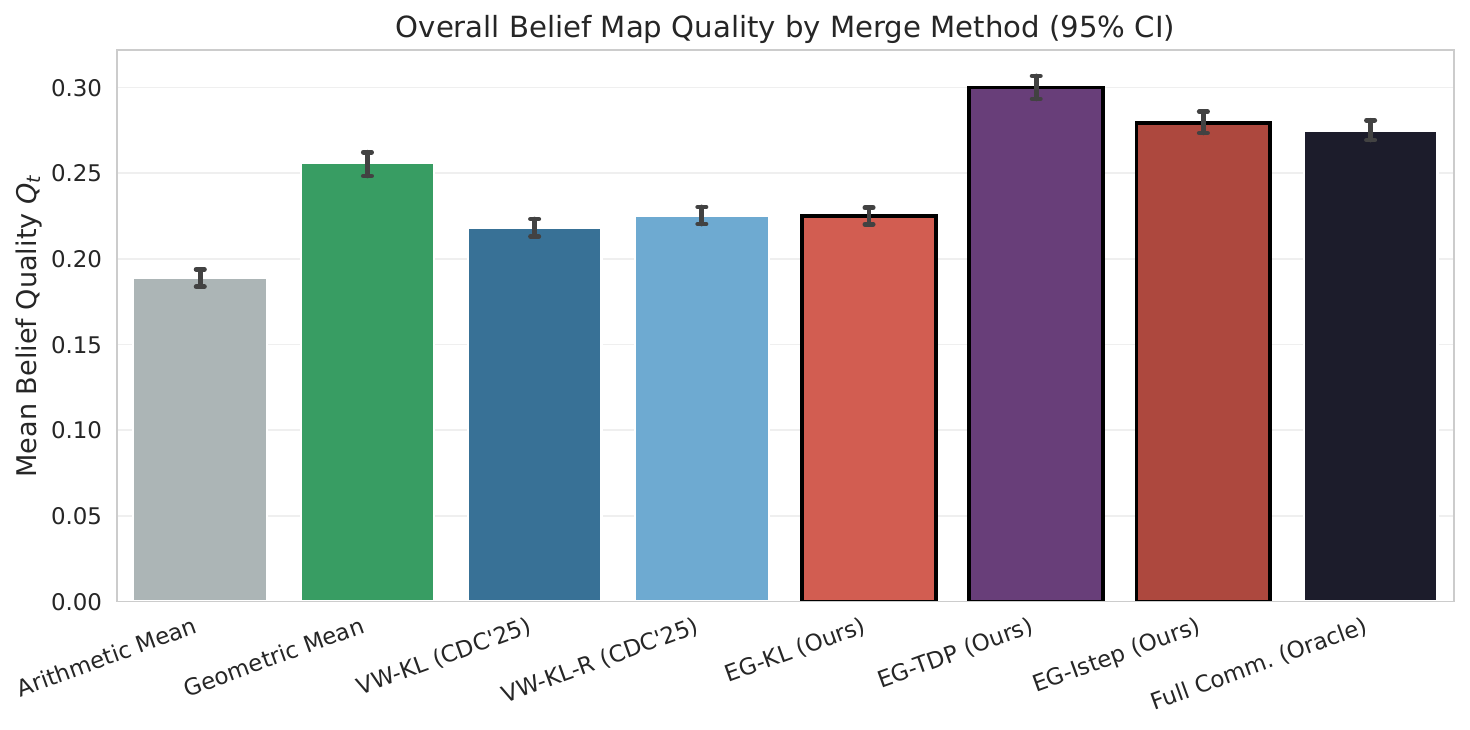}
  \figsep
  \caption{Overall \(\bar{Q}\) per method (95\% CI, 12{,}960 trials per method).
    VW-KL-R is shown for completeness, results are nearly identical to
    EG-KL and it is omitted from Table~\ref{tab:summary}.}
  \label{fig:summary}
\end{figure}

Figure~\ref{fig:summary} and Table~\ref{tab:summary} report \(\bar{Q}\) over all 103{,}680 trials. \textbf{EG-TDP} achieves the highest overall \(\bar{Q}{=}0.300\), a \(\mathbf{58.9\%}\) gain over \textsc{Arith} and \(\mathbf{37.4\%}\) over \textsc{VW-KL}, while simultaneously achieving the highest success rate (\(88.2\%\)) and shortest discovery time (317~steps on successful trials). \textbf{EG-KL} scores \(\bar{Q}{=}0.225\) (\(+19.2\%\) over \textsc{Arith}), however, this difference is not statistically significant against \textsc{VW-KL} (Welch $t$-test $p{=}0.058$, two-proportion $z$-test on success rate $p{=}0.158$), suggesting the gate alone does not substantially improve over the visit-weighted baseline. The primary gain from gating is enabling the planner switch in \textbf{EG-TDP}. \textbf{EG-KL}'s $3.2\%$ improvement in $\bar{Q}$ over \textsc{VW-KL} is not statistically significant ($p{=}0.058$), while \textbf{EG-TDP} adds a further $33.3\%$ over \textbf{EG-KL} (Table~\ref{tab:summary}). \textsc{Geom} achieves \(\bar{Q}{=}0.256\) but at the lowest success rate (\(79.2\%\)) because geometric pooling assigns zero probability to any cell where any agent's belief is zero. Once the target moves to a cleared region, no subsequent observations can restore mass at that cell.

\begin{table}[t]
\centering
\caption{Overall \(\bar{Q}\) and discovery statistics
  (\(\alpha{=}0.1\), \(\beta{=}0.2\), Poisson comm., 12{,}960
  trials per method). Steps on successful trials only.
  \label{tab:summary}}
\begin{tabular}{@{}lccc@{}}
\toprule
Method & \(\bar{Q}\;\pm\;\sigma\) & Succ. & Steps \\
\midrule
\textsc{Arith}  & $0.189\pm0.278$ & $83.0\%$ & $431$ \\
\textsc{Geom}   & $0.256\pm0.405$ & $79.2\%$ & $389$ \\
\textsc{VW-KL}  & $0.218\pm0.296$ & $86.2\%$ & $430$ \\
\midrule
\textbf{EG-KL}  & $0.225\pm0.298$ & $86.8\%$ & $434$ \\
\textbf{EG-TDP} & $\mathbf{0.300}\pm0.401$ & $\mathbf{88.2\%}$ & $\mathbf{317}$ \\
\bottomrule
\multicolumn{4}{@{}l@{}}{\scriptsize Ablations:
  TD-KL $\bar{Q}{=}0.021$, TD-KL-A $\bar{Q}{=}0.033$
  (Sec.~\ref{sec:ablation}).}
\end{tabular}
\end{table}

\subsection{Scaling with Grid Size}

\begin{figure}[t]
  \centering
  \includegraphics[width=\columnwidth]{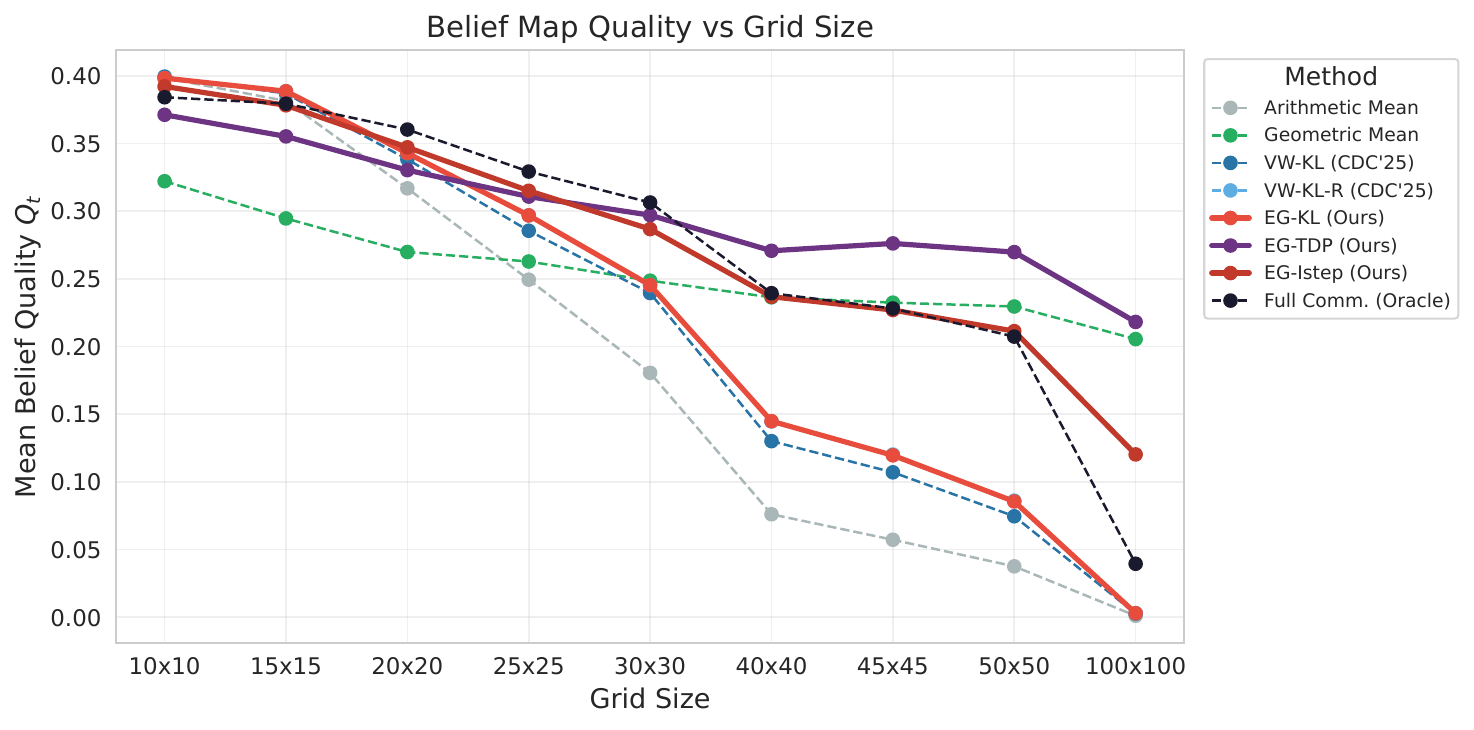}
  \figsep
  \caption{\(\bar{Q}\) vs grid size.
    \textbf{EG-KL} (red solid line) leads at \(10{\times}10\) but falls
    to \(\bar{Q}{\approx}0.003\) at \(100{\times}100\).
    \textbf{EG-TDP} (purple solid line) maintains \(\bar{Q}{=}0.218\) at the
    largest scale, validating the planner switch.
    \textsc{Geom} degrades gracefully but at the cost of the lowest success rate.}
  \label{fig:gridsize}
\end{figure}

Figure~\ref{fig:gridsize} shows \(\bar{Q}\) across all nine grid sizes. At \(10{\times}10\), \textbf{EG-KL} (\(\bar{Q}{=}0.398\)) leads \textbf{EG-TDP} (\(\bar{Q}{=}0.371\)). At small scale, entropy-reduction MPC alone is sufficient to converge on the target. As grid size grows, the two methods diverge sharply. At \(100{\times}100\), \textbf{EG-TDP} maintains \(\bar{Q}{=}0.218\) (54\% success) while \textbf{EG-KL} collapses to \(\bar{Q}{\approx}0.003\) (45\% success), a gap of over \(\mathbf{70{\times}}\). \textsc{VW-KL} (\(\bar{Q}{=}0.003\)) and \textsc{Arith} (\(\bar{Q}{=}0.001\)) also collapse at this scale. Only \textbf{EG-TDP} and \textsc{Geom} (\(\bar{Q}{=}0.206\)) maintain meaningful quality at \(100{\times}100\). The failure of EG-KL to scale despite entropy gating demonstrates that the planner switch, and not gating alone, enables convergence within a fixed step budget at large scale. The $70{\times}$ $\bar{Q}$ gap does not translate proportionally to success rate (45\% vs 54\%) because at this scale success depends primarily on spatial coverage. In other words, for large grids with not enough step budget, if we found the target was the agents acting based on informative belief, or did they lose track of where the target is. This is where $\bar{Q}$ differentiate what success rate alone cannot, even if methods have similar success rates.

\subsection{Discovery Rate and Steps}

\begin{figure}[t]
  \centering
  \includegraphics[width=\columnwidth]{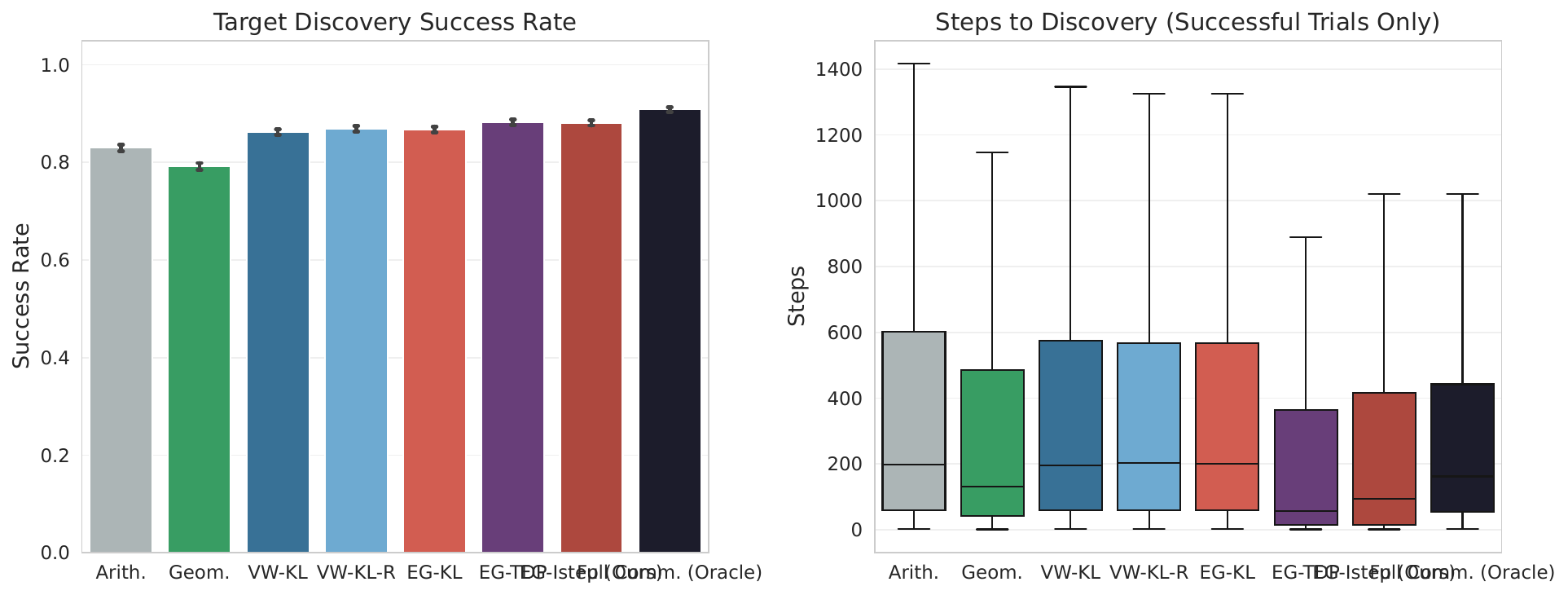}
  \figsep
  \caption{Success rate (left) and steps to discovery on successful trials (right).
    \textbf{EG-TDP} achieves the highest success rate and fastest discovery simultaneously.
    \textsc{Geom} finds the target quickly when it succeeds (389~steps) but fails most often (20.8\% failure rate).
    All pairwise success-rate differences are statistically significant at $p{<}0.001$ (two-proportion $z$-test, Bonferroni-corrected) except \textsc{VW-KL} vs \textbf{EG-KL} ($p{=}0.158$)}
  \label{fig:binary}
\end{figure}

Figure~\ref{fig:binary} confirms that \textbf{EG-TDP} exceeds on all three primary metrics. The steps axis covers successful trials only, since including the 2500-step penalty for failures would inflate every method's average and obscure the speed difference. The detection-probability planner switch drives agents toward the high-probability region as soon as the team has sufficient collective confidence, explaining both the higher success rate and shorter discovery time.

\subsection{Communication Interval Robustness}

\textbf{EG-TDP} maintains nearly flat \(\bar{Q}\) across all tested intervals (\(\bar{k}\in\{5,\ldots,200\}\). \(\bar{Q}\) ranges from 0.301 at \(\bar{k}{=}5\) to 0.293 at \(\bar{k}{=}200\). \textbf{EG-KL} is similarly flat (0.229 to 0.221). Notably, \textsc{Geom} and \textsc{Arith} slightly \emph{increase} with longer intervals (0.247\(\to\)0.266 and 0.179\(\to\)0.200). Infrequent merging gives individual agents more time to concentrate beliefs before averaging, incidentally reducing the harm of premature consensus. This is consistent with our entropy-gating argument.

\begin{figure}[t]
  \centering
  \includegraphics[width=0.5\textwidth]{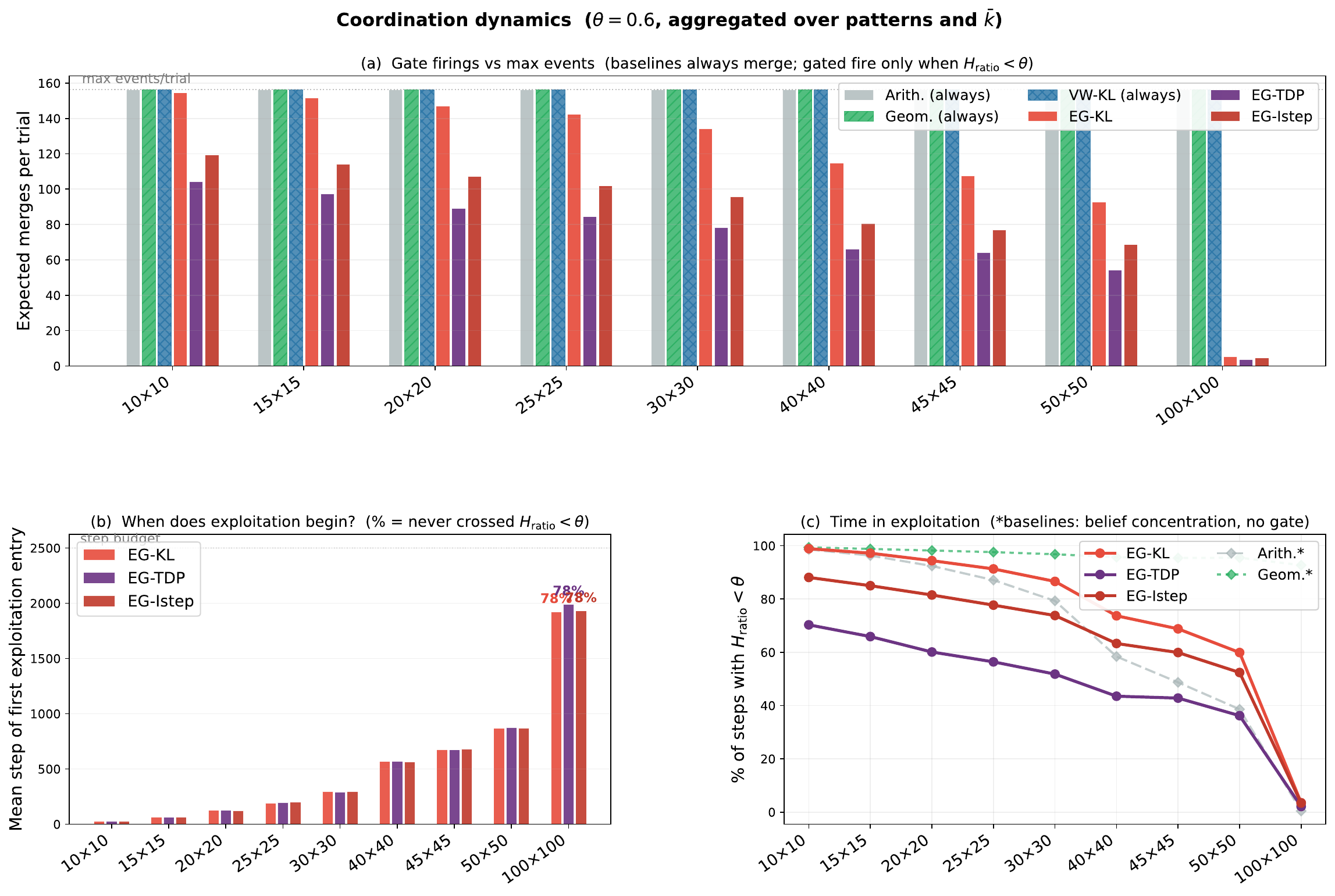}
  \figsep
  \caption{Coordination dynamics ($\theta{=}0.6$, aggregated over patterns and $\bar{k}$). \textbf{(a)}~Expected merges per trial (baselines always merge, EG-TDP fires only $\sim$3.6 of 156 events at $100{\times}100$, a 97\% bandwidth reduction). \textbf{(b)}~First exploitation step, annotations: \% of trials never crossing $\Hrat<\theta$. \textbf{(c)}~Fraction of steps in exploitation (for ungated methods, we show percentages of when entropy $< \theta$).}
  \label{fig:coordination_dynamics}
\end{figure}

The gate also provides a direct bandwidth benefit, at $100{\times}100$, \textbf{EG-TDP} fires only $\sim\!3.6$ merges per trial out of 156 expected communication events. Since each belief map at this scale is a 10{,}000-element vector, this reduces the volume of belief data exchanged per trial by over 97\% compared to unconditional fusion, without compromising discovery performance (Figure~\ref{fig:coordination_dynamics}a).
 
A natural question is whether performance saturates toward the communicate-always limit as \(\bar{k}\to0\). The \textsc{Full-Comm} reference (Table~\ref{tab:newmethods}), which updates a shared belief at every step, achieves \(\bar{Q}{=}0.552\) on the same representative configurations. \textbf{EG-TDP} at \(\bar{k}{=}5\) already exceeds this (\(\bar{Q}{=}0.593\)), confirming that the entropy gate provides a benefit absent in the communicate-always limit and that the method does not merely approach Full-Comm performance as communication becomes more frequent.

\subsection{Target Movement Pattern}\label{sec:results_pattern}

Table~\ref{tab:pattern} reports \(\bar{Q}\) and success rate by movement pattern. \textbf{EG-TDP} leads on stationary and patrol targets. Once a target exits the prior's support, all methods score \(\bar{Q}\leq0.095\), with \textsc{Geom} most affected by its zero-propagation recovery lag on patrol targets.
\begin{table}[t]
\centering
\caption{$\bar{Q}$ and success rate by target movement pattern.
  \label{tab:pattern}}
\begin{tabular}{@{}lcccc@{}}
\toprule
Method & Stationary & Patrol & Evasive & Random \\
\midrule
\textbf{EG-TDP} & \textbf{0.954} (100\%) & \textbf{0.226} (81\%) & 0.016 (80\%) & 0.003 (93\%) \\
\textbf{EG-KL}  & 0.579 (96\%)  & 0.224 (82\%) & 0.095 (80\%) & 0.003 (89\%) \\
\textsc{Geom}   & 0.926 (76\%)  & 0.077 (74\%) & 0.018 (75\%) & 0.003 (91\%) \\
\textsc{VW-KL}  & 0.563 (96\%)  & 0.213 (80\%) & 0.094 (79\%) & 0.003 (90\%) \\
\bottomrule
\end{tabular}
\end{table}

\subsection{Entropy Threshold Sensitivity}\label{sec:theta}

We swept \(\theta\in\{0.4,0.5,0.6,0.7,0.8\}\) on representative configurations (20$\times$20, 30$\times$30, 50$\times$50, stationary and patrol, 20~trials).

\begin{table}[t]
\centering
\caption{EG-TDP mean \(\bar{Q}\) by entropy threshold~\(\theta\) and
  expected communication interval~\(\bar{k}\).
  \label{tab:theta}}
\begin{tabular}{@{}ccccc@{}}
\toprule
\(\theta\) & \(\bar{k}{=}5\) & \(\bar{k}{=}25\) & \(\bar{k}{=}100\) \\
\midrule
0.4 & 0.460 & 0.477 & 0.469 \\
0.5 & 0.518 & 0.529 & 0.530 \\
0.6 & 0.594 & 0.599 & 0.586 \\
\textbf{0.7} & \textbf{0.602} & \textbf{0.602} & \textbf{0.596} \\
0.8 & 0.602 & 0.602 & 0.596 \\
\bottomrule
\end{tabular}
\end{table}

Table~\ref{tab:theta} shows two findings. First, performance peaks at \(\theta{=}0.7\) and is unchanged at \(\theta{=}0.8\). \textbf{EG-TDP} is insensitive to \(\theta\) above 0.7, so the threshold does not require precise tuning. Second, the optimal \(\theta\) is the same across all three tested communication intervals, so no interval-adaptive formulation is needed. \textbf{EG-KL} is flat across all \(\theta\) values (0.411--0.413). The reason this differs from \textbf{EG-TDP}'s sensitivity below \(\theta{=}0.7\) is left for future investigation, but the pattern suggests the planner switch is necessary for the threshold to have meaningful effect. Our main experiments used \(\theta{=}0.6\), which is slightly conservative, \(\theta{=}0.7\) yields an additional \(\sim\!1.5\%\) improvement.

\subsection{Proposed Method Comparison}\label{sec:newmethods}

Table~\ref{tab:newmethods} compares all three proposed methods against the unconditional baselines and the \textsc{Full-Comm} reference on representative configurations (stationary + patrol, three grid sizes, 20 trials each). Absolute \(\bar{Q}\) values are higher than Table~\ref{tab:summary} because these configurations exclude evasive and random patterns.

\textbf{EG-TDP vs Full-Comm.} \textbf{EG-TDP} achieves \(\bar{Q}{=}0.593\) versus 0.552 for \textsc{Full-Comm}. \textbf{EG-KL} (\(\bar{Q}{=}0.412\)) uses the same entropy gate but no planner switch, and scores below \textsc{Full-Comm}. The planner switch is therefore necessary, entropy gating alone is insufficient to outperform an always-communicating baseline on these configurations. A full account of how gated merging and the planner switch interact is left for future work.

\textbf{EG-TDP vs EG-Istep.} \textsc{EG-Istep} (\(\bar{Q}{=}0.543\)) scores below \textbf{EG-TDP} (0.593) but above \textsc{Full-Comm} (0.552). For our sensors (\(\Istep{=}0.114\)~bits), \(\gamma\) collapses to $\approx\!10^{-10}$ after 200 steps, reverting \textsc{EG-Istep} to near-pure exploration even at \(\bar{k}{=}5\). \textbf{EG-TDP}'s hard threshold is better calibrated at low \(\Istep\). For higher-\(\Istep\) sensors where \(\gamma\) decays slowly, \textsc{EG-Istep} may be preferable.

\begin{table}[t]
\centering
\caption{Proposed method comparison on representative configurations
  (stationary + patrol, three grid sizes, 20~trials, \(\theta{=}0.6\)).
  \label{tab:newmethods}}
\begin{tabular}{@{}lccc@{}}
\toprule
Method & \(\bar{Q}\;\pm\;\sigma\) & Succ. & Steps \\
\midrule
\textsc{Arith}              & $0.313\pm0.284$ & $89.7\%$ & $517$ \\
\textsc{VW-KL}              & $0.396\pm0.299$ & $91.1\%$ & $421$ \\
\textbf{EG-KL}              & $0.412\pm0.294$ & $92.6\%$ & $434$ \\
\textbf{EG-Istep}           & $0.543\pm0.379$ & $92.9\%$ & $319$ \\
Full-Comm (reference)       & $0.552\pm0.318$ & $95.4\%$ & $323$ \\
\textbf{EG-TDP}             & $\mathbf{0.593}\pm0.395$ & $\mathbf{93.8\%}$ & $\mathbf{296}$ \\
\bottomrule
\multicolumn{4}{@{}l@{}}{\scriptsize Full-Comm produces exact joint
  posterior; all merge-based methods approximate it.}
\end{tabular}
\end{table}

\subsection{Why Trust Decay in the Belief Map Fails}\label{sec:ablation}

Applying \(\gamma\) directly as \(\bmerged{=}\gamma\bvisit+(1{-}\gamma)\bprior\) reduces \(\bar{Q}\) to 0.021 (\textsc{TD-KL}) and 0.033 (\textsc{TD-KL-A}), reductions of 83\% and 74\% below \textsc{Arith}. The failure is structural, \(\gamma\) is identical for a correct peaked belief and a ghost because both produce the same entropy. Moving the trust signal to the planner objective (\textbf{EG-TDP}) avoids this entirely.

\section{Conclusion}

We introduced entropy-gated belief coordination for decentralized multi-agent target search under intermittent communication. The entropy ratio $\Hrat$ signals the exploration--exploitation transition. Below threshold $\theta$ the team merges beliefs and, in \textsc{EG-TDP}, switches to detection-probability planning. We derived $\Istep(\alpha,\beta)$ as a communication-free measure of per-step belief divergence. Across 103{,}680 trials, \textsc{EG-TDP} achieves $\bar{Q}{=}0.300$ ($+58.9\%$ over arithmetic mean), scales to $100{\times}100$ grids where unconditional methods collapse, and outperforms a joint-Bayesian reference on representative configurations. Future work includes partial communication graphs, heterogeneous per-agent $\Istep$, and continuous sensor models.

\bibliographystyle{IEEEtran}

\end{document}